\documentclass{article} 
\usepackage{iclr2023_conference_tinypaper,times}

\usepackage{amsmath,amsfonts,bm}

\def\eqref#1{equation~\ref{#1}}

\def\1{\bm{1}}

\DeclareMathAlphabet{\mathsfit}{\encodingdefault}{\sfdefault}{m}{sl}
\SetMathAlphabet{\mathsfit}{bold}{\encodingdefault}{\sfdefault}{bx}{n}

\usepackage{hyperref}
\usepackage{url}

\title{Generating Counterfactual Explanations Using Cardinality Constraints}

\author{Rub\'en Ruiz-Torrubiano \\
Institute of Digitalization and Informatics \\
IMC Krems University of Appplied Sciences\\
Krems, 3500, Austria \\
\texttt{ruben.ruiz@fh-krems.ac.at} \\
}

\iclrfinalcopy 
\begin{document}

\maketitle

\begin{abstract}
Providing explanations about how machine learning algorithms work and/or make 
particular predictions is one of the main tools that can be used to improve their 
trusworthiness, fairness and robustness. Among the most intuitive type of explanations 
are counterfactuals, which are examples that differ from a given point only in the 
prediction target and some set of features, presenting which features need to be
changed in the original example to flip the prediction for that example. However, such 
counterfactuals can have many different features than the original example, making their 
interpretation difficult. In this paper,
we propose to explicitly add a cardinality constraint to counterfactual generation 
limiting how many features can be different from the original example, thus providing more 
interpretable and easily understantable counterfactuals. 
\end{abstract}

\section{Introduction}

Explainable Artificial Intelligence (XAI) can be defined as the study and implementation of 
methods than provide visibility into how an AI system makes decisions, predictions and executes its 
actions \citep{rai_explainable_2020}. In general, two principal dimensions can be defined 
to classify XAI methods: whether the method requires knowledge of the model being explained, 
and whether the explanations refer to the model itself or its predictions \citep{du_techniques_2019}.\\ 
In the first case, 
knowing the internal workings of a particular algorithm results in \emph{model-specific} approaches,
whereas those
that handle machine learning models basically as black-boxes and can therefore be applied to a more 
general class of algorithms are called \emph{model-agnostic}. Examples of model-specific approaches 
are DNN-specific methods like those proposed in \citep{simonyan_deep_2014, fond_interpretable_2017, du_dnn_2018}.
By contrast, model-agnostic explanations use approaches like perturbations to determine feature contributions 
based on how sensitive the prediction target reacts when changing those features \citep{robnik-sikonja_perturbation-based_2018, ribeiro_why_2016, liu_nlize_2019}. 
Conterfactual explanations \citep{wachter_counterfactual_2017} can be considered to belong to this type of explanations.
The main idea is the following: let's consider a machine learning model $f_\theta$ and an input data point 
$\mathbf{x}$. We want to find data points $\hat{\mathbf{x}}$ such that they are the closest points to $\mathbf{x}$ such that the prediction 
target is different $f_\theta(\mathbf{x}) \neq f_\theta(\hat{\mathbf{x}})$. These data points should help the user understand 
what features one would need to change in $\mathbf{x}$ to flip the prediction target. For instance, let
$\mathbf{x}$ represent a loan application, and $f_\theta$ a model trained to predict if the loan will be paid off 
or will result in default. Having a prediction of default would likely result in the loan application being 
rejected. A counterfactual example $\hat{\mathbf{x}}$ would provide an explanation on which features would need to have 
been different in order to reach the opposite decision (e.g. by reducing the loan amount 
or having a higher income). \\
One of the main obstacles for using counterfactual explanations in practice is the amount of features that 
are different in the original example and the counterfactual: the higher, the more complicated and unintuitive 
counterfactual explanations can be. Even if a counterfactual is close to the original example in feature space
(say, in terms of the Euclidean distance between $\mathbf{x}$ and $\hat{\mathbf{x}}$), slight changes in a high number of features
can have a negative effect on its interpretability. Therefore, we propose in this work to add 
\emph{cardinality constraints} to counterfactual generation methods in order to ensure that the explanations 
provided do not diverge from the original example by more than $k$ features. Specifically, we provide an 
extension for the CERTIFAI framework \citep{sharma_certifai_2020} with cardinality constraints to answer the question 
of whether such \emph{sparse} counterfactuals can be generated effectively and efficiently. In general, there is a trade-off 
between sparse counterfactuals and those that minimize other quantitative measures like diversity or proximity 
\citep{mothilal_explaining_2020}. In this study, we focus on sparsity as our main criterium and leave using sparsity 
in combination with other measures as future work.  

\section{Methodology}
\label{section_methodology}

The CERTIFAI framework uses a custom genetic algorithm to find those counterfactuals $\hat{\mathbf{x}}$ that minimize
the distance $d(\mathbf{x}, \hat{\mathbf{x}})$ to a given data point $\mathbf{\mathbf{x}}$. In this paper, we restrict ourselves to tabular data and 
use as distance function the sum of the the $L_1$ distance (for continuous features) and a 
matching distance for categorial values as proposed in \cite{sharma_certifai_2020}. We forked the 
publicly available repository from the authors\footnote{\url{https://github.com/Ighina/CERTIFAI}} and 
implemented an additional cardinality constraint by penalizing those individuals with a cardinality 
(number of modified features with respect to the input example) higher than the target value $k$\footnote{\url{https://github.com/IMC-UAS-Krems/CERTIFAI_card}}. 
We compare the best counterfactuals generated by CERTIFAI without cardinality constraints (which are distance-based) 
with our counterfactuals and analyze their interpretability using random examples. To have a better overview,
we also calculate the mean cardinality between the constrained and unconstrained 
counterfactuals for each training example in the dataset used (see next section). We provide more details on the 
implementation as well as links to our code in Appendix \ref{appendix_a}. 

\section{Results and Discussion}

We performed experiments using the drug200 dataset (obtained from Kaggle)
to generate one counterfactual for each training example with and without cardinality constraints. 
In this dataset, there are in total 5 features, both continuous and categorical. Using this dataset, CERTIFAI computed 
counterfactuals with an average cardinality of $\hat{k}=3.1$.
We set the cardinality constraint 
to $k=2$ and $k=3$ features to get counterfactuals that are as easy as possible to interpret 
and compare the generated counterfactuals with the unconstrained ones. Table \ref{cf_comparison} shows a comparison of
counterfactuals obtained for a random example. As can be seen in the second row, the unconstrained counterfactuals 
generated by CERTIFAI can be different than the original sample in many features. Interpreting such a counterfactual 
might be difficult, as the only feature that stays with the same value in this case is the gender. By contrast, the low-cardinality 
counterfactuals generated with our approach are more easily interpretable: in the case $k=2$ the counterfactual can be 
intepreted as 'the target would change if the age is 15 and the NaToK ratio increases to 22.92'. The 
last row shows a counterfactual constrained to have a maximum of $k=3$ different features, which is also more easily 
interpretable than the unconstrained counterfactual. We provide additional experiments with another dataset in Appendix \ref{appendix_b}
to further support our results. 

\begin{table}[t]
    \caption{Comparison of counterfactuals obtained without cardinality constraints and with a cardinality constraint of $k=2$
        and $k=3$. Different values compared to the original example are shown in bold.}
    \label{cf_comparison}
    \begin{center}
    \begin{tabular}{rrrrrr}
    & Age & Sex & BP & Cholesterol & NaToK \\
    \hline
    Original & 16 & M & LOW & HIGH & 12.006 \\
    Unconstrained & \textbf{17} & M & \textbf{NORMAL} & \textbf{NORMAL} & \textbf{11.29} \\
    $k=2$ & \textbf{15} & M & LOW & HIGH & \textbf{22.82} \\
    $k=3$ & \textbf{15} & M & \textbf{HIGH} & HIGH & \textbf{11.04} \\
    \end{tabular}
    \end{center}
\end{table}

\section{Conclusions}

In this paper, we have presented a modification of the CERTIFAI framework to obtain low-cardinality counterfactuals as 
model-agnostic explanations. The presented results show that the cardinality-constrained counterfactuals are more easily interpretable.
As future work, we plan to design more effective genetic operators and validate our approach with larger datasets.

\subsubsection*{URM Statement}
The authors acknowledge that at least one key author of this work meets the URM criteria of ICLR 2024 Tiny Papers Track.

\bibliography{iclr2023_conference_tinypaper}

\begin{thebibliography}{11}
\providecommand{\natexlab}[1]{#1}
\providecommand{\url}[1]{\texttt{#1}}
\expandafter\ifx\csname urlstyle\endcsname\relax
  \providecommand{\doi}[1]{doi: #1}\else
  \providecommand{\doi}{doi: \begingroup \urlstyle{rm}\Url}\fi

\bibitem[Du et~al.(2018)Du, Liu, Song, and Hu]{du_dnn_2018}
Mengnan Du, Ninghao Liu, Qingquan Song, and Xia Hu.
\newblock Towards explanation of dnn-based prediction with guided feature
  inversion.
\newblock In \emph{Proceedings of the 24th ACM SIGKDD International Conference
  on Knowledge Discovery \& Data Mining}, KDD '18, pp.\  1358–1367, New York,
  NY, USA, 2018. Association for Computing Machinery.
\newblock ISBN 9781450355520.
\newblock \doi{10.1145/3219819.3220099}.
\newblock URL \url{https://doi.org/10.1145/3219819.3220099}.

\bibitem[Du et~al.(2019)Du, Liu, and Hu]{du_techniques_2019}
Mengnan Du, Ninghao Liu, and Xia Hu.
\newblock Techniques for interpretable machine learning.
\newblock \emph{Communications of the ACM}, 63\penalty0 (1):\penalty0 68--77,
  2019.
\newblock ISSN 0001-0782.
\newblock \doi{10.1145/3359786}.
\newblock URL \url{https://doi.org/10.1145/3359786}.

\bibitem[Fong \& Vedaldi(2017)Fong and Vedaldi]{fond_interpretable_2017}
Ruth~C. Fong and Andrea Vedaldi.
\newblock Interpretable explanations of black boxes by meaningful perturbation.
\newblock In \emph{2017 IEEE International Conference on Computer Vision
  (ICCV)}, pp.\  3449--3457, 2017.
\newblock \doi{10.1109/ICCV.2017.371}.

\bibitem[Liu et~al.(2019)Liu, Li, Li, Srikumar, Pascucci, and
  Bremer]{liu_nlize_2019}
Shusen Liu, Zhimin Li, Tao Li, Vivek Srikumar, Valerio Pascucci, and Peer-Timo
  Bremer.
\newblock {NLIZE}: {A} {Perturbation}-{Driven} {Visual} {Interrogation} {Tool}
  for {Analyzing} and {Interpreting} {Natural} {Language} {Inference} {Models}.
\newblock \emph{IEEE Transactions on Visualization and Computer Graphics},
  25\penalty0 (1):\penalty0 651--660, January 2019.
\newblock ISSN 1941-0506.
\newblock \doi{10.1109/TVCG.2018.2865230}.
\newblock URL \url{https://ieeexplore.ieee.org/document/8454904}.
\newblock Conference Name: IEEE Transactions on Visualization and Computer
  Graphics.

\bibitem[Mothilal et~al.(2020)Mothilal, Sharma, and
  Tan]{mothilal_explaining_2020}
Ramaravind~K. Mothilal, Amit Sharma, and Chenhao Tan.
\newblock Explaining machine learning classifiers through diverse
  counterfactual explanations.
\newblock In \emph{Proceedings of the 2020 {Conference} on {Fairness},
  {Accountability}, and {Transparency}}, {FAT}* '20, pp.\  607--617, New York,
  NY, USA, January 2020. Association for Computing Machinery.
\newblock ISBN 978-1-4503-6936-7.
\newblock \doi{10.1145/3351095.3372850}.
\newblock URL \url{https://doi.org/10.1145/3351095.3372850}.

\bibitem[Rai(2020)]{rai_explainable_2020}
Arun Rai.
\newblock Explainable {AI}: from black box to glass box.
\newblock \emph{Journal of the Academy of Marketing Science}, 48\penalty0
  (1):\penalty0 137--141, January 2020.
\newblock ISSN 1552-7824.
\newblock \doi{10.1007/s11747-019-00710-5}.
\newblock URL \url{https://doi.org/10.1007/s11747-019-00710-5}.

\bibitem[Ribeiro et~al.(2016)Ribeiro, Singh, and Guestrin]{ribeiro_why_2016}
Marco~Tulio Ribeiro, Sameer Singh, and Carlos Guestrin.
\newblock "{Why} {Should} {I} {Trust} {You}?": {Explaining} the {Predictions}
  of {Any} {Classifier}.
\newblock In \emph{Proceedings of the 22nd {ACM} {SIGKDD} {International}
  {Conference} on {Knowledge} {Discovery} and {Data} {Mining}}, {KDD} '16, pp.\
   1135--1144, New York, NY, USA, August 2016. Association for Computing
  Machinery.
\newblock ISBN 978-1-4503-4232-2.
\newblock \doi{10.1145/2939672.2939778}.
\newblock URL \url{https://dl.acm.org/doi/10.1145/2939672.2939778}.

\bibitem[Robnik-Šikonja \& Bohanec(2018)Robnik-Šikonja and
  Bohanec]{robnik-sikonja_perturbation-based_2018}
Marko Robnik-Šikonja and Marko Bohanec.
\newblock Perturbation-{Based} {Explanations} of {Prediction} {Models}.
\newblock In Jianlong Zhou and Fang Chen (eds.), \emph{Human and {Machine}
  {Learning}: {Visible}, {Explainable}, {Trustworthy} and {Transparent}},
  Human–{Computer} {Interaction} {Series}, pp.\  159--175. Springer
  International Publishing, Cham, 2018.
\newblock ISBN 978-3-319-90403-0.
\newblock \doi{10.1007/978-3-319-90403-0_9}.
\newblock URL \url{https://doi.org/10.1007/978-3-319-90403-0_9}.

\bibitem[Sharma et~al.(2020)Sharma, Henderson, and Ghosh]{sharma_certifai_2020}
Shubham Sharma, Jette Henderson, and Joydeep Ghosh.
\newblock {CERTIFAI}: {A} {Common} {Framework} to {Provide} {Explanations} and
  {Analyse} the {Fairness} and {Robustness} of {Black}-box {Models}.
\newblock In \emph{Proceedings of the {AAAI}/{ACM} {Conference} on {AI},
  {Ethics}, and {Society}}, {AIES} '20, pp.\  166--172, New York, NY, USA,
  February 2020. Association for Computing Machinery.
\newblock ISBN 978-1-4503-7110-0.
\newblock \doi{10.1145/3375627.3375812}.
\newblock URL \url{https://doi.org/10.1145/3375627.3375812}.

\bibitem[Simonyan et~al.(2014)Simonyan, Vedaldi, and
  Zisserman]{simonyan_deep_2014}
Karen Simonyan, Andrea Vedaldi, and Andrew Zisserman.
\newblock Deep inside convolutional networks: Visualising image classification
  models and saliency maps.
\newblock In Yoshua Bengio and Yann LeCun (eds.), \emph{2nd International
  Conference on Learning Representations, {ICLR} 2014, Banff, AB, Canada, April
  14-16, 2014, Workshop Track Proceedings}, 2014.
\newblock URL \url{http://arxiv.org/abs/1312.6034}.

\bibitem[Wachter et~al.(2017)Wachter, Mittelstadt, and
  Russell]{wachter_counterfactual_2017}
Sandra Wachter, Brent~D. Mittelstadt, and Chris Russell.
\newblock Counterfactual explanations without opening the black box: Automated
  decisions and the {GDPR}.
\newblock \emph{CoRR}, abs/1711.00399, 2017.
\newblock URL \url{http://arxiv.org/abs/1711.00399}.

\end{thebibliography}
\bibliographystyle{iclr2023_conference_tinypaper}

\appendix
\section{Implementation Details}
\label{appendix_a}
As outlined in Section \ref{section_methodology}, we implemented the cardinality constraint using the publicly 
available Python code provided by the authors of \cite{sharma_certifai_2020}. We defined a new distance function 
\texttt{card\_distance} that calculates the amount of different features between a candidate counterfactual and
the original data point. This distance function provided a deviation $\Delta$ between the target maximum cardinality 
and the cardinality of the current counterfactual. A linear penalty term was then calculated using a coefficient 
$c_{card}$ that was chosen in such a way that it effectively resulted in the cardinality constraint being considered 
as a hard constraint (i.e. $c_{card}\Delta>>\max fitness$). The original distance-based objective function used in CERTIFAI 
remained unchanged (i.e. the best counterfactuals found by CERTIFAI are scored according to this function). As a base model, we used the original model provided 
in the code which is a multi-layer perceptron with a hidden layer of $h=25$ neurons. We note that the results obtained 
are independent of the base model used, as our method is model-agnostic. The genetic algorithm is run 
for 10 generations and the probability of mutation and crossover are set to $p_m=0.2$ and $p_c=0.5$.

\section{Additional Experiments}
\label{appendix_b}
In order to further validate our results, we performed additional experiments using the Car Evaluation dataset 
(obtained from the UCI Machine Learning repository\footnote{\url{https://archive.ics.uci.edu/dataset/19/car+evaluation}}).
This dataset contains six categorial features, where the target shows the acceptance level of the car according to 
the features (buying price, maintenance price, number of doors, capacity in terms of persons to carry, the size 
of the luggage boot and an estimated level of safety). The dataset comprises 1728 rows. We run our algorithm to find 
counterfactuals with a maximum of $k=3$ different features. In this case, the average cardinality was $\hat{k}=1.6$, 
showing that our algorithm was able to find counterfactuals which are often of cardinality $k=2$ or even $k=1$.
Table \ref{cf_comparison_appendix} shows an additional comparison between an unconstrained counterfactual calculated 
by CERTIFAI and sparse counterfactuals calculated with our method. As can be seen, using the cardinality constraint 
can lead to sparser counterfactuals were changing already only very few features can already lead to flip the 
prediction target. 

\begin{table}[!ht]
    \caption{Comparison of counterfactuals obtained without cardinality constraints and with a cardinality constraint of $k=2$
        and $k=3$. Different values compared to the original example are shown in bold.}
    \label{cf_comparison_appendix}
    \begin{center}
    \begin{tabular}{rrrrrrr}
    & Buying & Maint & Doors & Persons & Lug\_boot & Safety \\
    \hline
    Original & vhigh & med & 2 & 2 & small & med \\
    Unconstrained & \textbf{low} & \textbf{low}  & 2 & \textbf{more} & small & \textbf{high} \\
    $k=2$ & vhigh & med & 2 & \textbf{more} & \textbf{big} & med \\
    $k=3$ & vhigh & med & \textbf{4} & \textbf{more} & \textbf{big} & med \\
    \end{tabular}
    \end{center}
\end{table}
\end{document}